\documentclass[11pt,a4paper]{article}
\usepackage[hyperref]{naaclhlt2019}
\usepackage{times}
\usepackage{latexsym}

\usepackage{booktabs}
\usepackage{footnote}

\usepackage{amsmath,amssymb,multicol,mathrsfs}
\usepackage[ruled,linesnumbered]{algorithm2e}
\usepackage{graphicx}
\usepackage{balance}

\usepackage{epstopdf}
\usepackage{multirow}
\usepackage{url}
\usepackage[skip=0pt]{subcaption}
\usepackage{color,soul}

\usepackage{tabularx}
 
\usepackage[normalem]{ulem}
\usepackage{enumitem}
\setlist{leftmargin=5.5mm}

\usepackage{url}

\aclfinalcopy % Uncomment this line for the final submission
\def\aclpaperid{35} %  Enter the acl Paper ID here

\title{LeafNATS: An Open-Source Toolkit and Live Demo System \\ for Neural Abstractive Text Summarization}

\author{Tian Shi \\
  Virginia Tech \\
  {\tt tshi@vt.edu} \\\And
  Ping Wang \\
  Virginia Tech \\
  {\tt ping@vt.edu} \\\And
  Chandan K. Reddy \\
  Virginia Tech \\
  {\tt reddy@cs.vt.edu}}

\date{}

\begin{document}
\maketitle
\begin{abstract}

Neural abstractive text summarization (NATS) has received a lot of attention in the past few years from both industry and academia.
In this paper, we introduce an open-source toolkit, namely LeafNATS, for training and evaluation of different sequence-to-sequence based models for the NATS task, and for deploying the pre-trained models to real-world applications.
The toolkit is modularized and extensible in addition to maintaining competitive performance in the NATS task.
A live news blogging system has also been implemented to demonstrate how these models can aid blog/news editors by providing them suggestions of headlines and summaries of their articles.

\end{abstract}

\section{Introduction}

Being one of the prominent natural language generation tasks, neural abstractive text summarization (NATS) has gained a lot of popularity~\cite{rush2015neural,see2017get,paulus2017deep}.
Different from extractive text summarization~\cite{gambhir2017recent, nallapati2017summarunner, verma2017extractive}, NATS relies on modern deep learning models, particularly sequence-to-sequence (Seq2Seq) models, to generate words from a vocabulary based on the representations/features of source documents~\cite{rush2015neural,nallapati2016abstractive}, 
so that it has the ability to generate high-quality summaries that are verbally innovative and can also easily incorporate external knowledge~\cite{see2017get}.
Many NATS models have achieved better performance in terms of the commonly used evaluation measures (such as \textit{ROUGE}~\cite{lin2004rouge} score) compared to extractive text summarization approaches~\cite{paulus2017deep,celikyilmaz2018deep,gehrmann2018bottom}.

We recently provided a comprehensive survey of the Seq2Seq models~\cite{shi2018neural}, including their network structures, parameter inference methods, and decoding/generation approaches, for the task of abstractive text summarization. A variety of NATS models share many common properties and some of the key techniques are widely used to produce well-formed and human-readable summaries that are inferred from source articles, such as encoder-decoder framework~\cite{sutskever2014sequence}, word embeddings~\cite{mikolov2013distributed}, attention mechanism~\cite{bahdanau2014neural}, pointing mechanism~\cite{vinyals2015pointer} and beam-search algorithm~\cite{rush2015neural}.
Many of these features have also found applications in other language generation tasks, such as machine translation~\cite{bahdanau2014neural} and dialog systems~\cite{serban2016building}.
In addition, other techniques that can also be shared across different tasks include training strategies~\cite{goodfellow2014generative,keneshloo2018deep,ranzato2015sequence}, data pre-processing, results post-processing and model evaluation.
Therefore, having an open-source toolbox that modularizes different network components and unifies the learning framework for each training strategy can benefit researchers in language generation from various aspects, including efficiently implementing new models and generalizing existing models to different tasks.

In the past few years, different toolkits have been developed to achieve this goal.
Some of them were designed specifically for a single task, such as ParlAI~\cite{miller2017parlai} for dialog research, and some have been further extended to other tasks.
For example, OpenNMT~\cite{klein2017opennmt} and XNMT~\cite{neubig2018xnmt} are primarily for neural machine translation (NMT), but have been applied to other areas.
The bottom-up attention model \cite{gehrmann2018bottom}, which has achieved state-of-the-art performance for abstractive text summarization, is implemented in OpenNMT.
There are also several other general purpose language generation packages, such as 
Texar~\cite{hu2018texar}.
Compared with these toolkits, LeafNATS is specifically designed for NATS research, but can also be adapted to other tasks.
In this toolkit, we implement an end-to-end training framework that can minimize the effort in writing codes for training/evaluation procedures, so that users can focus on building models and pipelines.
This framework also makes it easier for the users to transfer pre-trained parameters of user-specified modules to newly built models.

In addition to the learning framework, we have also developed a web application, which is driven by databases, web services and NATS models, to show a demo of deploying a new NATS idea to a real-life application using LeafNATS. 
Such an application can help front-end users (e.g., blog/news authors and editors) by providing suggestions of headlines and summaries for their articles.

The rest of this paper is organized as follows:
Section~\ref{sec:leafnats} introduces the structure and design of LeafNATS learning framework.
In Section~\ref{sec:demo}, we describe the architecture of the live system demo.
Based on the request of the system, we propose and implement a new model using LeafNATS for headline and summary generation.
We conclude this paper in Section~\ref{sec:conclusion}.

\section{LeafNATS Toolkit\footnote{\url{https://github.com/tshi04/LeafNATS}}}
\label{sec:leafnats}

\begin{figure}[!tp]
	\centering
	\includegraphics[width=0.38\textwidth]{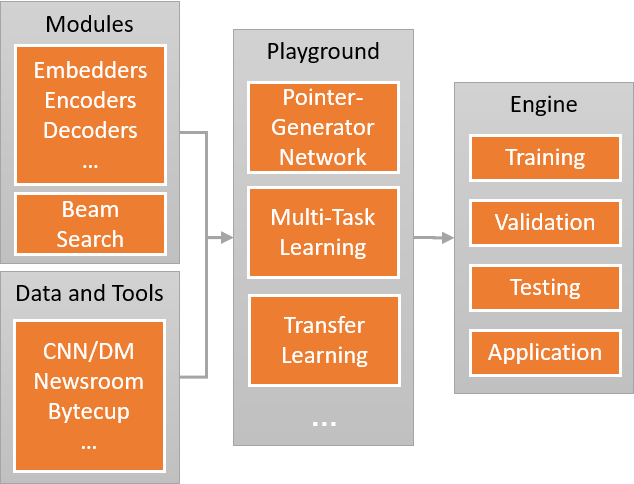}
	\vspace{-2mm}
	\caption{The framework of LeafNATS toolkit.}
	\label{fig:arthi}
	\vspace{-6mm}
\end{figure}

In this section, we introduce the structure and design of LeafNATS toolkit, which is built upon the lower level deep learning platform -- Pytorch~\cite{paszke2017automatic}.
As shown in Fig.~\ref{fig:arthi}, it consists of four main components, i.e., engines, modules, data and tools and playground.

\textbf{Engines:}
In LeafNATS, an engine represents a training algorithm.
For example, end-to-end training~\cite{see2017get} and adversarial training~\cite{goodfellow2014generative} are two different training frameworks. Therefore, we need to develop two different engines for them.

Specifically for LeafNATS, we implement a task-independent end-to-end training engine for NATS, but it can also be adapted to other NLP tasks, such as NMT, question-answering, sentiment classification, etc.
The engine uses abstract data, models, pipelines, and loss functions to build procedures of training, validation, testing/evaluation and application, respectively, so that they can be completely reused when implementing a new model.
For example, these procedures include saving/loading check-point files during training, selecting N-best models during validation, and using the best model for generation during testing, etc.
Another feature of this engine is that it allows users to specify part of a neural network to train and reuse parameters from other models, which is convenient for transfer learning.

\textbf{Modules:}
Modules are the basic building blocks of different models.
In LeafNATS, we provide ready-to-use modules for constructing recurrent neural network (RNN)-based sequence-to-sequence (Seq2Seq) models for NATS, e.g., pointer-generator network \cite{see2017get}.
These modules include embedder, RNN encoder, attention \cite{luong2015effective}, temporal attention \cite{nallapati2016abstractive}, attention on decoder \cite{paulus2017deep} and others.
We also use these basic modules to assemble a pointer-generator decoder module and the corresponding beam search algorithms.
The embedder can also be used to realize the embedding-weights sharing mechanism~\cite{paulus2017deep}.

\textbf{Data and Tools:}
Different models in LeafNATS are tested on three datasets (see Table~\ref{tab:data}), namely, CNN/Daily Mail (CNN/DM)~\cite{hermann2015teaching}, Newsroom~\cite{grusky2018newsroom} and Bytecup\footnote{\url{https://biendata.com/competition/bytecup2018/}}.
The pre-processed CNN/DM data is available online\footnote{\url{https://github.com/JafferWilson/Process-Data-of-CNN-DailyMail}}.
Here, we provide tools to pre-process the last two datasets.
Data modules are used to prepare the input data for mini-batch optimization.

\begin{table}[htp]
	\renewcommand\arraystretch{1.}
	\centering
	\resizebox{1.\linewidth}{!}{
	\begin{tabular}{|>{\centering\arraybackslash}p{6em}|>{\centering\arraybackslash}p{5em}|>{\centering\arraybackslash}p{5em}|>{\centering\arraybackslash}p{5em}|}
		\hline
		 \bf Dataset & \bf Train & \bf Validation & \bf Test \\
		\hline
		CNN/DM & 287,227 & 13,368 & 11,490 \\\hline
		Newsroom & 992,985 & 108,612 & 108,655\\\hline
		Bytecup & 892,734 & 111,592 & 111,592\\\hline
	\end{tabular}}
	\vspace{-3mm}
	\caption{Basic statistics of the datasets used.}
	\vspace{-4mm}
    \label{tab:data}
\end{table}

\textbf{Playground:}
With the engine and modules, we can develop different models by just assembling these modules and building pipelines in playground.
We re-implement different models in the NATS toolkit~\cite{shi2018neural} to this framework.
The performance (ROUGE scores \cite{lin2004rouge}) of the pointer-generator model on different datasets has been reported in Table~\ref{tab:perform}, where we find that most of the results are better than our previous implementations~\cite{shi2018neural} due to some minor changes to the neural network.

\begin{table}[htp]
	\renewcommand\arraystretch{1.}
	\centering
	\resizebox{1.\linewidth}{!}{
	\begin{tabular}{|c|c|c|c|c|}
		\hline
		\bf Dataset & \bf Model &\bf R-1 & \bf R-2 & \bf R-L \\\hline
		Newsroom-S & Pointer-Generator & 39.91 & 28.38 & 36.87 \\\hline
		Newsroom-H & Pointer-Generator & 27.11 & 12.48 & 25.47 \\\hline
		\multirow{2}[0]{*}{CNN/DM} & Pointer-Generator & 37.02 & 15.97 & 34.18 \\\cline{2-5}
		& +coverage & 39.26 & 17.21 & 36.16 \\\hline
		Bytecup & Pointer-Generator & 40.50 & 24.57 & 37.63\\\hline
	\end{tabular}}
	\caption{Performance of our implemented pointer-generator network on different datasets. Newsroom-S and -H represent Newsroom summary and headline datasets, respectively.}
    \label{tab:perform}
    \vspace{-5mm}
\end{table}

\section{A Live System Demonstration\footnote{\url{http://dmkdt3.cs.vt.edu/leafNATS}}}
\label{sec:demo}

In this section, we present a real-world web application of the abstractive text summarization models, which can help front-end users to write headlines and summaries for their articles/posts.
We will first discuss the architecture of the system, and then, provide more details of the front-end design and a new model built by LeafNATS that makes automatic summarization and headline generation possible.

\subsection{Architecture}

\begin{figure}[!tp]
	\centering
	\includegraphics[width=0.42\textwidth]{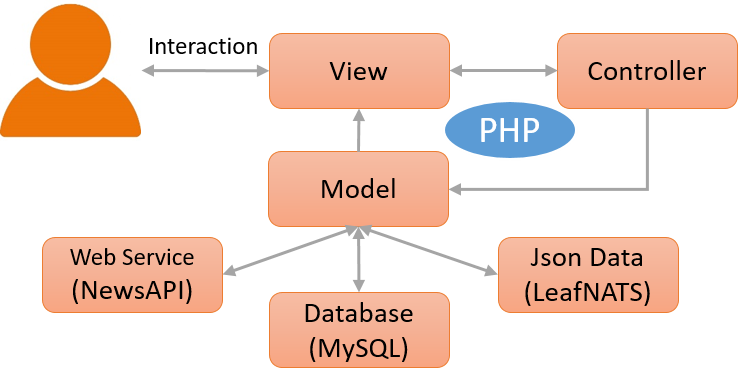}
	\vspace{-2mm}
	\caption{The architecture of the live system.}
	\vspace{-5mm}
	\label{fig:mvc}
\end{figure}

This is a news/blog website, which allows people to read, duplicate, edit, post, delete and comment articles.
It is driven by web-services, databases and our NATS models.
This web application is developed with PHP, HTML/CSS, and jQuery following the concept of Model-View-Controller (see Fig.~\ref{fig:mvc}).

In this framework, when people interact with the front-end views, they send HTML requests to controllers that can manipulate models. 
Then, the views will be changed with the updated information.
For example, in NATS, we first write an article in a text-area.
Then, this article along with the summarization request will be sent to the controller via jQuery Ajax call.
The controller communicates with our NATS models asynchronously via JSON format data.
Finally, generated headlines and summaries are shown in the view.

\subsection{Design of Frontend}

\begin{figure*}[!tp]
	\centering
	\includegraphics[width=\textwidth]{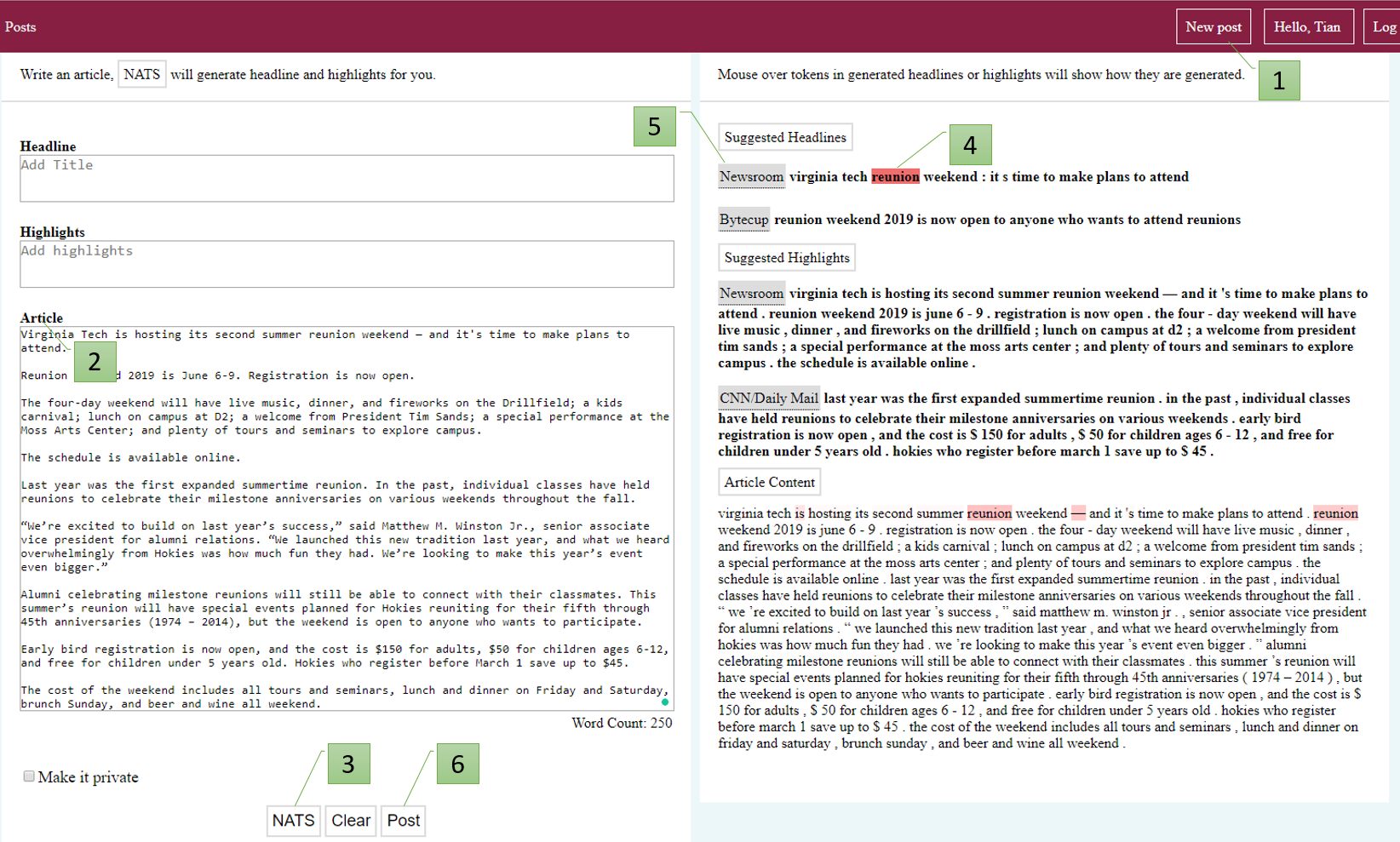}
	\caption{Front-end design of the live demonstration of our system.}
	\label{fig:frontend}
	\vspace{-5mm}
\end{figure*}

Fig.~\ref{fig:model} presents the front-end design of our web application for creating a new post, where labels represent the sequence of actions.
In this website, an author can first click on ``New Post'' (step 1) to bring a new post view.
Then, he/she can write content of an article in the corresponding text-area (step 2) without specifying it's headline and highlights, i.e., summary.
By clicking ``NATS'' button (step 3) and waiting for a few seconds, he/she will see the generated headlines and highlights for the article in a new tab on the right hand side of the screen.
Here, each of the buttons in gray color denotes the resource of the training data.
For example, ``Bytecup'' means the model is trained with Bytecup headline generation dataset.
The tokenized article content is shown in the bottom.
Apart from plain-text headlines and highlights, our system also enables users to get a visual understanding of how each word is generated via attention weights~\cite{luong2015effective}.
When placing the mouse tracker (step 4) on any token in the headlines or highlights, related content in the article will be labeled with red color.
If the author would like to use one of the suggestions, he/she can click on the gray button (step 5) to add it to the text-area on the left hand side and edit it.
Finally, he/she can click ``Post'' (step 6) to post the article.

\subsection{The Proposed Model}

\begin{figure}[!tp]
	\centering
	\includegraphics[width=0.42\textwidth]{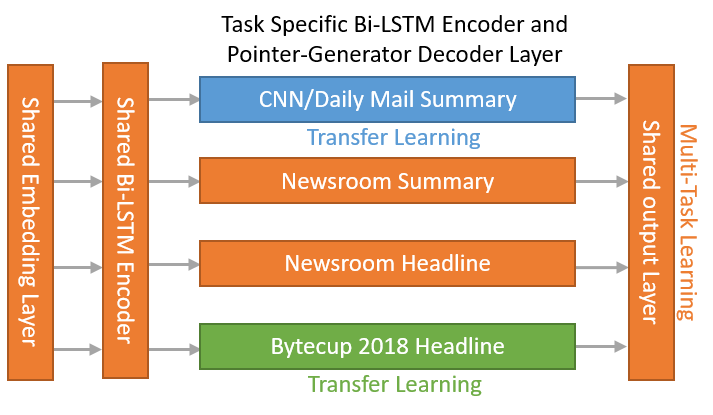}
	\vspace{-2mm}
	\caption{Overview of the model used to generate headlines and summaries.}
	\label{fig:model}
	\vspace{-5mm}
\end{figure}

As shown in the Fig.~\ref{fig:frontend}, our system can suggest to the users two headlines (based on Newsroom headline and Bytecup datasets) and summaries (based on Newsroom summary and CNN/DM datasets).
They are treated as four tasks in this section.
To achieve this goal, we use the modules provided in LeafNATS toolkit to assemble a new model (see Fig.~\ref{fig:model}), which has a shared embedding layer, a shared encoder layer, a task specific encoder-decoder (Bi-LSTM encoder and pointer-generator decoder) layer and a shared output layer. 

To train this model, we first build a multi-task learning pipeline for Newsroom dataset to learn parameters for the modules that are colored in orange in Fig.~\ref{fig:model}, because (1) articles in this dataset have both headlines and highlights, (2) the size of the dataset is large, and (3) the articles come from a variety of news agents.
Then, we build a transfer learning pipeline for CNN/Daily and Bytecup dataset, and learn the parameters for modules labeled with blue and green color, respectively.
With LeafNATS, we can accomplish this work efficiently.

The performance of the proposed model on the corresponding testing sets are shown in Table~\ref{tab:perform2}.
From the table, we observe that our model performs better in headline generation tasks.
However, the ROUGE scores in summarization tasks are lower than the models without sharing embedding, encoder and output layers.
It should be noted that by sharing the parameters, this model requires less than 20 million parameters to achieve such performance.

\begin{table}[!tp]
	\renewcommand\arraystretch{1.}
	\centering
	\resizebox{1.\linewidth}{!}{
	\begin{tabular}{|c|>{\centering\arraybackslash}p{6em}|c|c|c|}
		\hline
		\bf Dataset & \bf Model &\bf R-1 & \bf R-2 & \bf R-L \\\hline
		Newsroom-S & multi-task & 39.85 & 28.37 & 36.91 \\\hline
		Newsroom-H & multi-task & 28.31 & 13.40 & 26.64  \\\hline
		\multirow{2}[0]{*}{CNN/DM} & transfer & 35.55 & 15.19 & 33.00\\\cline{2-5}
		& +coverage & 38.49 & 16.78 & 35.68 \\\hline
		Bytecup & transfer & 40.92 & 24.51 & 38.01\\\hline
	\end{tabular}}
	\vspace{-2mm}
	\caption{Performance of our model.}
    \label{tab:perform2}
    \vspace{-5mm}
\end{table}

\section{Conclusion}
\label{sec:conclusion}

In this paper, we have introduced a LeafNATS toolkit for building, training, testing/evaluating, and deploying NATS models, as well as a live news blogging system to demonstrate how the NATS models can make the work of writing headlines and summaries for news articles more efficient.
An extensive set of experiments on different benchmark datasets has demonstrated the effectiveness of our implementations.
The newly proposed model for this system has achieved competitive results with fewer number of parameters.

\section*{Acknowledgments}
This work was supported in part by the US National Science Foundation grants IIS-1619028, IIS-1707498 and IIS-1838730.

\bibliography{naaclhlt2019}

\begin{thebibliography}{27}
\expandafter\ifx\csname natexlab\endcsname\relax\def\natexlab#1{#1}\fi

\bibitem[{Bahdanau et~al.(2014)Bahdanau, Cho, and Bengio}]{bahdanau2014neural}
Dzmitry Bahdanau, Kyunghyun Cho, and Yoshua Bengio. 2014.
\newblock Neural machine translation by jointly learning to align and
  translate.
\newblock \emph{arXiv preprint arXiv:1409.0473}.

\bibitem[{Celikyilmaz et~al.(2018)Celikyilmaz, Bosselut, He, and
  Choi}]{celikyilmaz2018deep}
Asli Celikyilmaz, Antoine Bosselut, Xiaodong He, and Yejin Choi. 2018.
\newblock Deep communicating agents for abstractive summarization.
\newblock In \emph{Proceedings of the 2018 Conference of the North American
  Chapter of the Association for Computational Linguistics: Human Language
  Technologies, Volume 1 (Long Papers)}, volume~1, pages 1662--1675.

\bibitem[{Gambhir and Gupta(2017)}]{gambhir2017recent}
Mahak Gambhir and Vishal Gupta. 2017.
\newblock Recent automatic text summarization techniques: a survey.
\newblock \emph{Artificial Intelligence Review}, 47(1):1--66.

\bibitem[{Gehrmann et~al.(2018)Gehrmann, Deng, and Rush}]{gehrmann2018bottom}
Sebastian Gehrmann, Yuntian Deng, and Alexander Rush. 2018.
\newblock Bottom-up abstractive summarization.
\newblock In \emph{Proceedings of the 2018 Conference on Empirical Methods in
  Natural Language Processing}, pages 4098--4109.

\bibitem[{Goodfellow et~al.(2014)Goodfellow, Pouget-Abadie, Mirza, Xu,
  Warde-Farley, Ozair, Courville, and Bengio}]{goodfellow2014generative}
Ian Goodfellow, Jean Pouget-Abadie, Mehdi Mirza, Bing Xu, David Warde-Farley,
  Sherjil Ozair, Aaron Courville, and Yoshua Bengio. 2014.
\newblock Generative adversarial nets.
\newblock In \emph{Advances in neural information processing systems}, pages
  2672--2680.

\bibitem[{Grusky et~al.(2018)Grusky, Naaman, and Artzi}]{grusky2018newsroom}
Max Grusky, Mor Naaman, and Yoav Artzi. 2018.
\newblock Newsroom: A dataset of 1.3 million summaries with diverse extractive
  strategies.
\newblock In \emph{Proceedings of the 2018 Conference of the North American
  Chapter of the Association for Computational Linguistics: Human Language
  Technologies, Volume 1 (Long Papers)}, volume~1, pages 708--719.

\bibitem[{Hermann et~al.(2015)Hermann, Kocisky, Grefenstette, Espeholt, Kay,
  Suleyman, and Blunsom}]{hermann2015teaching}
Karl~Moritz Hermann, Tomas Kocisky, Edward Grefenstette, Lasse Espeholt, Will
  Kay, Mustafa Suleyman, and Phil Blunsom. 2015.
\newblock Teaching machines to read and comprehend.
\newblock In \emph{Advances in Neural Information Processing Systems}, pages
  1693--1701.

\bibitem[{Hu et~al.(2018)Hu, Shi, Yang, Tan, Zhao, He, Wang, Yu, Qin, Wang
  et~al.}]{hu2018texar}
Zhiting Hu, Haoran Shi, Zichao Yang, Bowen Tan, Tiancheng Zhao, Junxian He,
  Wentao Wang, Xingjiang Yu, Lianhui Qin, Di~Wang, et~al. 2018.
\newblock Texar: A modularized, versatile, and extensible toolkit for text
  generation.
\newblock \emph{arXiv preprint arXiv:1809.00794}.

\bibitem[{Keneshloo et~al.(2018)Keneshloo, Shi, Reddy, and
  Ramakrishnan}]{keneshloo2018deep}
Yaser Keneshloo, Tian Shi, Chandan~K Reddy, and Naren Ramakrishnan. 2018.
\newblock Deep reinforcement learning for sequence to sequence models.
\newblock \emph{arXiv preprint arXiv:1805.09461}.

\bibitem[{Klein et~al.(2017)Klein, Kim, Deng, Senellart, and
  Rush}]{klein2017opennmt}
Guillaume Klein, Yoon Kim, Yuntian Deng, Jean Senellart, and Alexander Rush.
  2017.
\newblock Opennmt: Open-source toolkit for neural machine translation.
\newblock \emph{Proceedings of ACL 2017, System Demonstrations}, pages 67--72.

\bibitem[{Lin(2004)}]{lin2004rouge}
Chin-Yew Lin. 2004.
\newblock {ROUGE}: A package for automatic evaluation of summaries.
\newblock \emph{Text Summarization Branches Out}.

\bibitem[{Luong et~al.(2015)Luong, Pham, and Manning}]{luong2015effective}
Thang Luong, Hieu Pham, and Christopher~D Manning. 2015.
\newblock Effective approaches to attention-based neural machine translation.
\newblock In \emph{Proceedings of the 2015 Conference on Empirical Methods in
  Natural Language Processing}, pages 1412--1421.

\bibitem[{Mikolov et~al.(2013)Mikolov, Sutskever, Chen, Corrado, and
  Dean}]{mikolov2013distributed}
Tomas Mikolov, Ilya Sutskever, Kai Chen, Greg~S Corrado, and Jeff Dean. 2013.
\newblock Distributed representations of words and phrases and their
  compositionality.
\newblock In \emph{Advances in neural information processing systems}, pages
  3111--3119.

\bibitem[{Miller et~al.(2017)Miller, Feng, Batra, Bordes, Fisch, Lu, Parikh,
  and Weston}]{miller2017parlai}
Alexander Miller, Will Feng, Dhruv Batra, Antoine Bordes, Adam Fisch, Jiasen
  Lu, Devi Parikh, and Jason Weston. 2017.
\newblock Parlai: A dialog research software platform.
\newblock In \emph{Proceedings of the 2017 Conference on Empirical Methods in
  Natural Language Processing: System Demonstrations}, pages 79--84.

\bibitem[{Nallapati et~al.(2017)Nallapati, Zhai, and
  Zhou}]{nallapati2017summarunner}
Ramesh Nallapati, Feifei Zhai, and Bowen Zhou. 2017.
\newblock Summarunner: A recurrent neural network based sequence model for
  extractive summarization of documents.
\newblock In \emph{Thirty-First AAAI Conference on Artificial Intelligence}.

\bibitem[{Nallapati et~al.(2016)Nallapati, Zhou, dos Santos, glar
  Gul{\c{c}}ehre, and Xiang}]{nallapati2016abstractive}
Ramesh Nallapati, Bowen Zhou, Cicero dos Santos, {\c{C}}a~glar Gul{\c{c}}ehre,
  and Bing Xiang. 2016.
\newblock Abstractive text summarization using sequence-to-sequence {RNNs} and
  beyond.
\newblock \emph{CoNLL 2016}, page 280.

\bibitem[{Neubig et~al.(2018)Neubig, Sperber, Wang, Felix, Matthews,
  Padmanabhan, Qi, Sachan, Arthur, Godard et~al.}]{neubig2018xnmt}
Graham Neubig, Matthias Sperber, Xinyi Wang, Matthieu Felix, Austin Matthews,
  Sarguna Padmanabhan, Ye~Qi, Devendra~Singh Sachan, Philip Arthur, Pierre
  Godard, et~al. 2018.
\newblock Xnmt: The extensible neural machine translation toolkit.
\newblock \emph{Vol. 1: MT Researchers’ Track}, page 185.

\bibitem[{Paszke et~al.(2017)Paszke, Gross, Chintala, Chanan, Yang, DeVito,
  Lin, Desmaison, Antiga, and Lerer}]{paszke2017automatic}
Adam Paszke, Sam Gross, Soumith Chintala, Gregory Chanan, Edward Yang, Zachary
  DeVito, Zeming Lin, Alban Desmaison, Luca Antiga, and Adam Lerer. 2017.
\newblock Automatic differentiation in pytorch.

\bibitem[{Paulus et~al.(2017)Paulus, Xiong, and Socher}]{paulus2017deep}
Romain Paulus, Caiming Xiong, and Richard Socher. 2017.
\newblock A deep reinforced model for abstractive summarization.
\newblock \emph{arXiv preprint arXiv:1705.04304}.

\bibitem[{Ranzato et~al.(2015)Ranzato, Chopra, Auli, and
  Zaremba}]{ranzato2015sequence}
Marc'Aurelio Ranzato, Sumit Chopra, Michael Auli, and Wojciech Zaremba. 2015.
\newblock Sequence level training with recurrent neural networks.
\newblock \emph{arXiv preprint arXiv:1511.06732}.

\bibitem[{Rush et~al.(2015)Rush, Chopra, and Weston}]{rush2015neural}
Alexander~M Rush, Sumit Chopra, and Jason Weston. 2015.
\newblock A neural attention model for abstractive sentence summarization.
\newblock In \emph{Proceedings of the 2015 Conference on Empirical Methods in
  Natural Language Processing}, pages 379--389.

\bibitem[{See et~al.(2017)See, Liu, and Manning}]{see2017get}
Abigail See, Peter~J. Liu, and Christopher~D. Manning. 2017.
\newblock Get to the point: Summarization with pointer-generator networks.
\newblock In \emph{Proceedings of the 55th Annual Meeting of the Association
  for Computational Linguistics (Volume 1: Long Papers)}, pages 1073--1083.
  Association for Computational Linguistics.

\bibitem[{Serban et~al.(2016)Serban, Sordoni, Bengio, Courville, and
  Pineau}]{serban2016building}
Iulian~Vlad Serban, Alessandro Sordoni, Yoshua Bengio, Aaron~C Courville, and
  Joelle Pineau. 2016.
\newblock Building end-to-end dialogue systems using generative hierarchical
  neural network models.

\bibitem[{Shi et~al.(2018)Shi, Keneshloo, Ramakrishnan, and
  Reddy}]{shi2018neural}
Tian Shi, Yaser Keneshloo, Naren Ramakrishnan, and Chandan~K Reddy. 2018.
\newblock Neural abstractive text summarization with sequence-to-sequence
  models.
\newblock \emph{arXiv preprint arXiv:1812.02303}.

\bibitem[{Sutskever et~al.(2014)Sutskever, Vinyals, and
  Le}]{sutskever2014sequence}
Ilya Sutskever, Oriol Vinyals, and Quoc~V Le. 2014.
\newblock Sequence to sequence learning with neural networks.
\newblock In \emph{Advances in neural information processing systems}, pages
  3104--3112.

\bibitem[{Verma and Lee(2017)}]{verma2017extractive}
Rakesh~M. Verma and Daniel Lee. 2017.
\newblock Extractive summarization: Limits, compression, generalized model and
  heuristics.
\newblock \emph{Computaci{\'o}n y Sistemas}, 21.

\bibitem[{Vinyals et~al.(2015)Vinyals, Fortunato, and
  Jaitly}]{vinyals2015pointer}
Oriol Vinyals, Meire Fortunato, and Navdeep Jaitly. 2015.
\newblock Pointer networks.
\newblock In \emph{Advances in Neural Information Processing Systems}, pages
  2692--2700.

\end{thebibliography}
\bibliographystyle{acl_natbib}

\end{document}